%% file: main.tex
\documentclass[journal]{IEEEtran}
\usepackage{amsmath,amsfonts}
\usepackage{algorithmic}
\usepackage{algorithm}
\usepackage{array}
\usepackage[caption=false,font=normalsize,labelfont=sf,textfont=sf]{subfig}
\usepackage{textcomp}
\usepackage{stfloats}
\usepackage{url}
\usepackage{verbatim}
\usepackage{graphicx}
\usepackage{cite}

\usepackage{amssymb}
\usepackage{makecell}
\usepackage{bm}
\usepackage{bbm}
\usepackage{arydshln}
\usepackage{booktabs}
\usepackage{hyperref}

\begin{document}

\title{HODN: Disentangling Human-Object Feature \\ for HOI Detection}


\author{
  Shuman Fang,
  Zhiwen Lin,
  Ke Yan,
  Jie Li,
  Xianming Lin$^*$,\\
  Rongrong Ji, \IEEEmembership{Senior Member,~IEEE}
\thanks{
    This work is done when S. Fang works as an intern at Youtu Laboratory, Tencent, Shanghai 200233, China.
}
\thanks{
    S. Fang, J. Li, X. Lin (Corresponding Author), and R. Ji are with the Media Analytics and Computing Laboratory, Department of Artificial Intelligence, School of Informatics, Xiamen University, Xiamen 361005, China. (fangshuman@stu.xmu.edu.cn, lijie.32@outlook.com, linxm@xmu.edu.cn, rrji@xmu.edu.cn)
}
\thanks{
    Z. Lin and K. Yan are with Youtu Laboratory, Tencent, Shanghai 200233, China. (xavier.lin@foxmail.com, kerwinyan@tencent.com)
}
\thanks{
    R. Ji is also with the Institute of Artificial Intelligence, Xiamen University, and Fujian Engineering Research Center of Trusted Artificial Intelligence Analysis and Application, Xiamen University, 361005, China. (rrji@xmu.edu.cn)
}
\thanks{
    $^*$Corresponding Author: Xianming Lin (linxm@xmu.edu.cn)
}
}



\maketitle

\input{sec_0_abstract.tex}
\begin{IEEEkeywords}
Human-Object Interaction Detection, Transformer, Visual Attention, Disentangling Features.
\end{IEEEkeywords}

\input{sec_1_introduction.tex}
\input{sec_2_related_work.tex}
\input{sec_3_method.tex}
\input{sec_4_experiment.tex}
\input{sec_5_conclusion.tex}
\input{acknowledgements.tex}

\bibliographystyle{IEEEtranS}
\bibliography{egbib}

\end{document}

%% file: sec_0_abstract.tex
\begin{abstract}
The task of Human-Object Interaction (HOI) detection is to detect humans and their interactions with surrounding objects,
where transformer-based methods show dominant advances currently.
However, these methods ignore the relationship among humans, objects, and interactions:
1) human features are more contributive than object ones to interaction prediction;
2) interactive information disturbs the detection of objects but helps human detection.
In this paper, we propose a \textit{Human and Object Disentangling Network} (HODN) to model the HOI relationships explicitly,
where humans and objects are first detected by two disentangling decoders independently and then processed by an interaction decoder.
Considering that human features are more contributive to interaction,
we propose a \textit{Human-Guide Linking} method to make sure
the interaction decoder focuses on the human-centric regions with human features as the positional embeddings.
To handle the opposite influences of interactions on humans and objects,
we propose a \textit{Stop-Gradient Mechanism} to stop interaction gradients from optimizing the object detection but to allow them to optimize the human detection.
%
Our proposed method achieves competitive performance on both the V-COCO and the HICO-Det datasets.
It can be combined with existing methods easily for state-of-the-art results.

\end{abstract}

%% file: sec_1_introduction.tex
\section{Introduction}
\IEEEPARstart{I}{nstance-level}
detection no longer satisfies the requirement of understanding the visual world, but the relationships inference among instances has attracted considerable research interest recently, where Human-Object Interaction (HOI) detection plays a major role.
In addition, other high-level semantic understanding tasks, such as activity recognition~\cite{wang2016context_actreg,xu2017hierarchical_actreg} and visual question answering~\cite{tsai2015study_vqa,liu2020visual_vqa}, can benefit from HOI.
The goal of HOI detection aims to detect humans and surrounding objects and infer the interactive relations between them,
which can be typically represented as triplets of $\langle$\textit{human, object, interaction}$\rangle$.
Hence, HOI detection consists of three parts: human detection, object detection, and interaction classification.

%
Based on variants of RCNN~\cite{ren2015faster,he2017mask}, conventional methods usually detect instances firstly and enumerate all human-object pairs to predict secondly~\cite{chao2018learning_two,gao2018ican_two,gkioxari2018detecting_two,li2020hoi_two,liu2020amplifying_two,hou2020visual_two,ulutan2020vsgnet_two,hou2021detecting_two,gao2020drg_two,li2019transferable_two,wan2019pose_two,zhong2020polysemy_two,li2020pastanet_two}, or directly identify the pairs that are likely to interact~\cite{liao2020ppdm_one,kim2020uniondet_one,wang2020learning_one,zhong2021glance_one}.
These methods
suffer from the lack of contextual information due to the locality of convolutional layers and pooling layers.
%
Nowadays, transformer-based HOI detectors~\cite{tamura2021qpic,zou2021end,kim2021hotr,chen2021reformulating,zhang2021mining} are proposed to handle this problem in an end-to-end manner.
Benefiting from the attention mechanism, these networks can extract global features rather than local ones.
%
%
%
However,
in the very beginning,
the transformer-based methods~\cite{tamura2021qpic,zou2021end} utilize a simple pipeline with a single encoder-decoder pair to detect the triplets of HOI,
which struggles with handling localization and classification simultaneously.
%
%
To deal with this problem, recent methods~\cite{kim2021hotr,chen2021reformulating,zhang2021mining} disentangle the HOI task as tasks of instance detection and interaction classification.
These methods usually utilize two decoders, whether parallel~\cite{zhang2021mining} or cascaded~\cite{kim2021hotr,chen2021reformulating}, to handle the corresponding tasks.
%
%
Among them,
DisTR~\cite{zhou2022human_distr} takes a further step to disentangle both encoders and decoders for the two sub-tasks.
%
%
The disentanglement of sub-tasks lets different modules focus on their corresponding tasks and improve the whole performance.
However,
for the instance detection task,
previous works regard a pair of human and object as one instance, and process the instance features as a whole,
which ignores the distinct effects between humans and objects.
%
%
We argue that humans and objects play different roles in HOI detection, and the mutual effect among humans, objects, and interactions should also be analyzed.

\input{Figures/compare.tex}

To dig out the comprehensive relationships among humans, objects, and interactions, we conduct experiments and analyze from two respects,
\emph{i.e.}, 1) how humans and objects impact interactions and 2) how interactions impact humans and objects.
In Section\,\ref{sec:exp_motivation},
we analyze these by masking the regions of different parts in the images and removing the module for interaction prediction.
The results show that:
1) both humans and objects make contributions to interaction prediction, but humans contribute much more;
2) human detection needs the help of interactions while object detection will be disturbed.

Motivated by the different effects between humans and objects, we emphasize the necessity of utilizing disentangled human-object features.
Hence, we propose an end-to-end transformer-based framework, termed \textit{Human and Object Disentangling Network} (HODN), to explicitly model the relationships among humans, objects, and interactions.
A simple comparison with previous works is visualized in Figure\,\ref{fig:comp}.
%
As depicted,
we use two separate detection decoders, \emph{i.e.}, \textit{human decoder} and \textit{object decoder}, to extract human and object features independently,
which will be processed by an \textit{interaction decoder}.
To make sure the human features contribute more there, a \textit{Human-Guide Linking} (HG-Linking) method is utilized for the interaction decoder to focus on the human-centric regions.
In particular, the interaction decoder receives the human features as the positional embeddings for all attention layers.
These position embeddings, \emph{a.k.a.}, interaction queries, are used to guide the decoder where to focus on.
To link humans with the surrounding objects,
object features are fed into the first layer of the interaction decoder to provide prior knowledge.
With this, the interaction decoder can not only make use of the information of humans and objects but also make the human features dominant and the object ones auxiliary.
Furthermore, due to another HOI relationship that interactive information obstructs object detection but helps human detection,
we propose a training strategy, termed \textit{Stop-Gradient Mechanism} (SG-Mechanism) to process interaction gradients separately.
During back-propagation, the SG-Mechanism stops interaction gradients from passing through the object decoder but maintains them into the human decoder the same as the common practice.
This not only inhibits the negative impact of interactions on object detection but keeps the slightly positive on human detection, which brings the best detection performance for both humans and objects.

To evaluate the performance of our method, we conduct extensive experiments by following the previous works on the widely used datasets, \emph{i.e.}, V-COCO\cite{gupta2015visual} and HICO-Det~\cite{chao2018learning_two}, where our method achieves competitive results.
%
Moreover, we combine two latest works~\cite{zhang2022exploring_stip,liao2022gen_genvlkt} with our method,
bringing 2.40\% and 0.91\% relative gains respectively, to achieve the state-of-the-art performance.
Visualization experiments also verified that our method can localize humans and objects more preciously and can focus on the interaction regions.
We owe these to the disentangling of different parts and the established comprehensive relationships among them.

Concretely, we summarize our work as:
1) we found that humans contribute more to interaction prediction, and interactions have opposite influences on the detection of humans and objects.
2) We propose HODN to explicitly model the relationships, where an HG-Linking is utilized by the interaction decoder to make human features dominant and object ones auxiliary, and an SG-Mechanism is proposed to handle interaction gradients differently.
The HODN achieves competitive experimental results and can be easily combined with existing methods for state-of-the-art performance.









%% file: Figures/compare.tex
\begin{figure*}[t]
    \centering
    \includegraphics[width=1\textwidth]{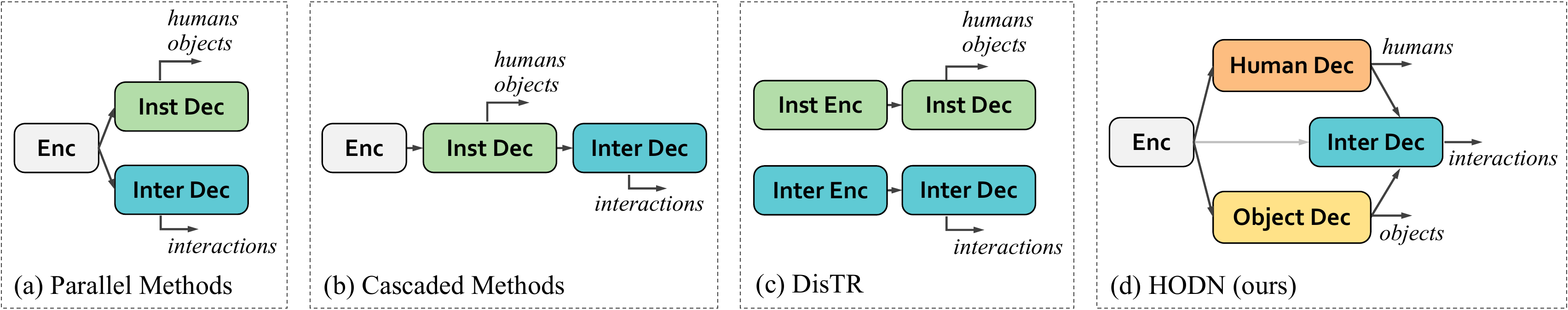}
    \caption{
        Comparison of recent disentangling transformer-based methods.
        Previous works usually disentangle the HOI task into instance detection and interaction prediction by introducing two decoders.
        (a) Parallel methods~\cite{kim2021hotr,chen2021reformulating} adopt two relatively independent decoders.
        (b) Cascaded methods~\cite{zhang2021mining} handle these parts in series.
        (c) DisTR~\cite{zhou2022human_distr} takes a further step based on parallel methods, where both encoder and decoder are disentangled.
        (d) Our HODN contains two parallel detection decoders for human and object features and one interaction prediction decoder.
        Enc: Encoder; Dec: Decoder; Inst: Instance; Inter: Interaction.
    }\label{fig:comp}
\end{figure*}

%% file: sec_2_related_work.tex
\section{Related Work}
Based on the used detector,
HOI methods can be categorized into two main streams: traditional methods and transformer-based methods, where the former are based on variants of R-CNN~\cite{ren2015faster,he2017mask} and the latter are on DETR~\cite{carion2020end}.

\subsection*{Traditional Methods}
Traditional methods can be further divided into two-stage and one-stage methods.
Two-stage methods~\cite{chao2018learning_two,gao2018ican_two,gkioxari2018detecting_two,li2020hoi_two} rely on an off-the-shelf object detector to localize all instances, including humans and objects.
Then they enumerate all human-object combinations, crop the features inside the localized regions, and process cropped features by multi-stream networks.
Typically, the multi-stream networks include a human stream, an object stream, and a pairwise stream.
To improve the performance, some works introduce extra streams with extra knowledge, including spatial features~\cite{liu2020amplifying_two,hou2020visual_two,ulutan2020vsgnet_two}, word embeddings~\cite{hou2021detecting_two,gao2020drg_two}, human postures~\cite{li2019transferable_two,wan2019pose_two,zhong2020polysemy_two,li2020pastanet_two}, or above in combination.
Another line of works also introduces graph neural networks to model the relationship of human-object pairs~\cite{yang2020graph_two,ulutan2020vsgnet_two,wang2020contextual_two,lin2021action_two,liu2020consnet_two}.
However, two-stage methods suffer from finding the interactive human-object pairs in an overwhelming number of permutations.

To alleviate the number of redundant non-interactive pairs,
one-stage methods introduce the concept of interaction point as the anchor to directly identify the pairs that are likely to interact~\cite{liao2020ppdm_one,kim2020uniondet_one,wang2020learning_one,zhong2021glance_one}.
By parallelly processing the instance detection branch and the interaction point prediction branch, they can match the most similar combination between the two independent branches to complete HOI detection.
In particular, the introduced interaction point is the midpoint of the center points of human and object~\cite{liao2020ppdm_one,liao2020ppdm_one}, or the center point of the human-object pair union box~\cite{kim2020uniondet_one}.
Instead of using one interaction point, GGNet~\cite{zhong2021glance_one} infers a set of points to predict interactions more robustly.
However, either one point or a set of points,
such a concept may introduce a lot of useless regions and even some misleading information,
\emph{i.e.},
one-stage methods may not work in situations
when the human and the object in the interaction are far apart or
when multiple instances are overlapping.

In summary, no matter the one-stage or the two-stage methods, traditional methods are highly dependent on the quality of object detection and can not capture the interactive relationships accurately due to the lack of contextual information in CNNs.

\subsection*{Transformer-based Methods}
Considering that transformer architectures have succeeded in many computer vision tasks~\cite{dosovitskiy2020image,touvron2021training},
recent works introduce the transformer into the HOI detection task based on DETR~\cite{carion2020end}.
Instead of relying on a pre-trained object detector, transformer-based methods generate a set of HOI triplets.
The attention mechanism in the transformer encoder can extract the global feature, including humans, objects, and context.
During the transformer decoder stage, these methods introduce a set of anchor-like learnable positional embeddings, \emph{a.k.a} queries, to predict HOI triplets directly in an end-to-end manner.
With the global feature, the network can easily mine the interactive information between humans and objects so that transformer-based methods can get quite advanced performance.
Moreover,
as the number of queries determines that of predicted results, transformer-based methods address the problem of a large number of redundant human-object pairs suffered by the two-stage ones.
And due to the learnable nature of queries, the regions where the network is interested can be optimized to solve the problem faced by the one-stage methods.

In the beginning, transformer-based works examine how to design the transformer decoders,
of which there exist two main streams, \emph{i.e.}, single-decoder and two-decoder methods.
The single-decoder methods, \emph{e.g.}, QPIC~\cite{tamura2021qpic} and HOI-Transformer~\cite{zou2021end},
use only one decoder to handle HOI detection.
The mixed features may make the decoder unable to focus on target regions.
Some works have realized the differences between attention regions of instances (the pairs of humans and objects) and interactions,
and then they disentangle the HOI task into two sub-tasks: \emph{i.e.}, instance detection and interaction prediction.
These two-decoder works can be further divided into cascaded methods~\cite{zhang2021mining} and parallel ones~\cite{kim2021hotr,chen2021reformulating}.
By providing instance features to guarantee the performance of interaction prediction,
the cascaded method CDN~\cite{zhang2021mining} detects instances firstly and predicts interactions secondly.
However, object detection will be disturbed by interactions due to the cascaded linking way.
To keep the instance decoder away from interaction impacts,
HOTR~\cite{kim2021hotr} and AS-Net~\cite{chen2021reformulating} use two parallel decoders to deal with instance detection and interaction prediction independently.
However, these methods still suffer from mis-grouped instance-interaction pairs.
Recent parallel work, DisTR~\cite{zhou2022human_distr}, utilizes an attention module to fuse instance features into interaction representations to provide joint configurations of them.
With the connection of instances and interactions,
DisTR proposes to further disentangle features, where the transformer encoder and decoders are decoupled.
%
%
%
%
Another work, HOD~\cite{zhang2023hod}, disentangles the decoder into three independent parts. 
Then, HOD introduces random erasing for the object decoder to improve generalization and pose information for the human decoder to augment representations.

No matter how the decoders are designed,
all existing methods ignore the differences between humans and objects,
\emph{i.e.}, they do not dig out the relationships among humans, objects, and interactions,
indicating that there exists much space for them to improve.

Some other transformer-based methods also try to introduce additional information to boost performance.
OCN~\cite{yuan2022detecting_transformer}, PhraseHOI~\cite{li2022improving_transformer} and GEN-VLKT~\cite{liao2022gen_genvlkt}
use word embeddings to assist in interaction prediction.
Besides linguistic features, STIP~\cite{zhang2022exploring_stip} also adopts spatial features with graph methods.
These methods also fail to explore the relationships among humans, objects, and interactions.
We argue that our method is flexible so that can combine with them easily to achieve higher performance.


%% file: sec_3_method.tex
\input{Figures/framework.tex}
\section{Method}
The overall architecture of our proposed HODN is shown in Figure\,\ref{fig:framework} and stated in Section\,\ref{sec:met_overall}.
Our framework is proposed to model the relationships among humans, objects, and interactions based on:
1) human features make more contributions to interaction prediction than object ones;
2) interactive information disturbs object detection but assists in human detection.
We propose a \textit{Human-Guide Linking method} (HG-Linking) in Section\,\ref{sec:met_on} to help the interaction decoder predict interactions by holding the former relationship.
In Section\,\ref{sec:met_from}, we introduce the detail of a training strategy named \textit{Stop-Gradient Mechanism} (SG-Mechanism), which is proposed to satisfy the latter one.

\subsection{Overall Architecture}
\label{sec:met_overall}
\subsubsection{Global Feature Extractor}
Given an image $\bm{x} \in {\mathbb{R}}^{3 \times H \times W}$, the CNN backbone extracts the visual feature map $\bm{z}_b \in {\mathbb{R}}^{C \times H' \times W'}$ from it,
where $H$ and $W$ are the height and width of the input image, $H'$ and $W'$ are those of the feature map, and $C$ is the number of channels.
Then the visual feature map $\bm{z}_b$ is reduced in channel dimension from $C$ to $d$ by a projection convolution layer with a kernel size of $1 \times 1$.
Next, the spatial dimensions of it are collapsed into one dimension by using a flatten operator.
The processed feature map $\bm{z}_{src} \in {\mathbb{R}}^{d \times (H' \times W')}$ combined with a positional encoding $\bm{p} \in {\mathbb{R}}^{d \times (H' \times W')}$ is fed into the transformer encoder
to get the global memory feature $\bm{z}_e$.
In this stage, the multi-head self-attention can focus on not only regions of humans and objects but also global contextual information.

\subsubsection{The HOI Decoders}
The global memory feature $\bm{z}_e$, along with the positional encoding $\boldsymbol{p}$, is then utilized by two parallel decoders to provide contextual information.
The two parallel decoders, the \textit{human decoder} and the \textit{object decoder}, are used to detect their corresponding targets independently.
The human decoder transforms a set of randomly initialized learnable positional embeddings
$\boldsymbol{Q}_H = \left\{
    \boldsymbol{q}^h_i
    \mid \boldsymbol{q}^h_i \in \mathbb{R}^d
\right\}_{i=1}^N$
(\textit{human queries}) into $\boldsymbol{Q}_H^{out}$ (\textit{human features}) layer by layer, where $d$ is the channel dimension of the encoder and $N$ is the number of positional embeddings.
So does the object decoder,
which transforms another set of learnable positional embeddings $\boldsymbol{Q}_O$ (\textit{object queries}) with the same size as $\boldsymbol{Q}_H$ into $\boldsymbol{Q}_O^{out}$ (\textit{object features}).
In this stage,
the human features and the object features with the same subscript are considered as the human-object pair automatically, \emph{i.e.}, the human-object pair features can be represented as
$\left\{
    \big(\boldsymbol{q}^{h,out}_i, \boldsymbol{q}^{o,out}_i \big)
    \mid \boldsymbol{q}^{h,out}_i \in \boldsymbol{Q}_H^{out},
         \boldsymbol{q}^{o,out}_i \in \boldsymbol{Q}_O^{out}
\right\}_{i=1}^N$.
Note that to guarantee this,
the two sets of queries, $\boldsymbol{Q}_H$ and $\boldsymbol{Q}_O$, are initialized to be equal.
Then, the \textit{interaction decoder} receives the human-object pair features,
\emph{i.e.}, $\boldsymbol{Q}_H^{out}$ and $\boldsymbol{Q}_O^{out}$,
to query possible human-object pairs and dig out interaction knowledge between them.
The output from the interaction decoder is denoted as $\boldsymbol{Q}_{A}^{out}$.

\subsubsection{Final Prediction Heads}
The final part of our framework includes four feed-forward networks (FFNs), \emph{i.e.}, a human-box FFN, an object-box FFN, an object-class FFN, and an interaction FFN.
The outputs of three decoders,
\emph{i.e.}, $\boldsymbol{Q}_H^{out}$, $\boldsymbol{Q}_O^{out}$, and $\boldsymbol{Q}_{A}^{out}$,
are then fed into the corresponding FFNs to get human-box vectors, object-box vectors, object-class vectors as well as interaction-class vectors.
These vectors share the same length in design, which is the same size of queries.
Therefore,
we can get a set of HOI predictions with the size of $N$, each of which is presented as
$\langle$\textit{human box, object box, object class, interaction class}$\rangle$.
With these HOI predictions,
we follow previous works~\cite{zou2021end,tamura2021qpic,kim2021hotr,zhang2021mining} for training and inference.

\subsection{Human-Guide Linking}
\label{sec:met_on}
We follow DETR~\cite{carion2020end} to design the human decoder and the object decoder.
However, unlike the human or object decoder which only processes one kind of information, the interaction decoder needs to fuse both human features and object features.
Considering that human features contribute more than objects,
we argue that naively taking the addition of two features as inputs will not bring satisfactory performance.
Hence, a specific link method between the interaction decoder and the others two decoders is non-trivial.

We first review the architectural details of the vanilla transformer decoder,
which takes a set of learnable positional embeddings, the global memory feature $\boldsymbol{z}_e$, and the positional encoding $\boldsymbol{p}$ as inputs to localize the bounding boxes and predict the classes of targets,
where the learnable positional embeddings, \emph{a.k.a.}, queries, are designed to learn the potential target regions.
The transformer decoder is a stack of decoder layers, each of which is composed of three main parts: a self-attention module, a cross-attention module, and a feed-forward network (FFN).
The self-attention module in the $i$-th layer can be formulated by the following:
\begin{align}
    \label{equ:self_attn}
    \boldsymbol{A}_{i}^{self} =
    \mathrm{softmax}
    \left(
        \frac
        {
            \big( \boldsymbol{Q}^{out}_{i-1} + \boldsymbol{Q} \big)
            \big( \boldsymbol{Q}^{out}_{i-1} + \boldsymbol{Q} \big)
            ^\mathsf{T}
        }
        {\sqrt{d}}
    \right) \boldsymbol{Q}^{out}_{i-1},
\end{align}
where $d$ denotes the channel dimension of the decoder, $\boldsymbol{Q}$ is the queries, and $\boldsymbol{Q}^{out}_{i-1}$ is the outputs of the previous decoder layer.
Note that the $\boldsymbol{Q}^{out}_{0}$ is initialized with zeros, so the self-attention module in the first layer is meaningless and can be skipped as mentioned in DETR~\cite{carion2020end}.
From Eq.\,\ref{equ:self_attn}, it can be seen that with the help of the queries, the self-attention module can capture the relationships among outputs of the previous layer and inhibit duplicate ones.
The cross-attention module in the $i$-th layer can be formulated as:
\begin{align}
    \label{equ:cross_attn}
    \boldsymbol{A}_{i}^{cross} =
    \mathrm{softmax}
    \left(
        \frac
        {
            \big( \boldsymbol{A}_{i}^{self} + \boldsymbol{Q}\big)
            \big( \boldsymbol{z}_e + \boldsymbol{p} \big)
            ^\mathsf{T}
        }
        {\sqrt{d}}
    \right) \boldsymbol{z}_e,
\end{align}
where $\boldsymbol{p}$ is the positional encoding used by the encoder and $\boldsymbol{z}_e$ is the global memory feature output from the encoder.
During the cross-attention module, $\boldsymbol{A}_{i}^{self}$ is aggregated with the highly responsive parts in $\boldsymbol{z}_e$ and is further refined to improve the detection of targets, where the queries provide information of the attention position.
Note that both in self-attention and cross-attention, the queries are utilized to supply the spatial information of the distinct regions.
Hence, we argue that the queries act as a guide to force the decoder where to focus on, which is also verified by DETR~\cite{carion2020end}.

\input{Figures/dec_layer.tex}

Different from the vanilla transformer decoder, including the human decoder and the object decoder, the interaction decoder needs to model the HOI relationships.
As analyzed before, there is a strong correlation between interactions and humans.
Hence,
based on the functionality of two attention modules, we propose a \textit{Human-Guide Linking} method to make human features more contributive.
The detail of this specific linking method is given in Figure\,\ref{fig:dec_layer}.
In particular,
we regard the human features $\boldsymbol{Q}_H^{out}$ from the human decoder as the positional embeddings, \emph{a.k.a.}, interaction queries, which are utilized by all attention layers to guide the interaction decoder where to concentrate on.
With this, the attention of the interaction decoder can be around humans.
However, just making the interaction decoder pay attention to human-centric regions is not sufficient.
Object features $\boldsymbol{Q}_O^{out}$ should also be considered albeit less contributive than human ones.
Instead of enumerating permutations that may generate $N \times N$ possible pairs, we consider a one-to-one same-subscript matching strategy to assign the human-object pairs.
So we regard the additions of $\boldsymbol{Q}_H^{out}$ and $\boldsymbol{Q}_O^{out}$ as assigned pairs and feed them into the self-attention module in the first layer to dig out the relation between them and remove duplication prediction.
Particularly, we change the attention formula of the self-attention in the first layer as below:
\begin{align}
    \label{equ:ho_attn}
    \boldsymbol{A}_{1}^{self} =
    \mathrm{softmax}
    \left(
        \frac
        {
            \big( \boldsymbol{Q}_H^{out} + \boldsymbol{Q}_O^{out} \big)
            \big( \boldsymbol{Q}_H^{out} + \boldsymbol{Q}_O^{out} \big)
            ^\mathsf{T}
        }
        {\sqrt{d}}
    \right) \boldsymbol{Q}_O^{out}.
\end{align}
Unlike the vanilla decoder where the self-attention in the first layer is meaningless, we make full use of it, which can construct the relationship between humans and objects quickly.
The attention modules in the following decoder layers only take human features into account, not object features anymore.
With human features dominant and object ones auxiliary,
the interaction decoder can effectively model the interaction relationships between humans and objects.

\subsection{Stop Gradient Mechanism}
\label{sec:met_from}
Considering another HOI relationship that interactions have a negative influence on object detection but a positive influence on human detection, we propose a special training strategy, \textit{Stop Gradient Mechanism} (SG-Mechanism), to handle the relationship.
As shown in our framework,
the outputs from the HOI Decoders are fed into the Final Prediction Heads,
\emph{i.e.},
the $\boldsymbol{Q}_{A}^{out}$ are sent into the interaction FFN,
the $\boldsymbol{Q}_H^{out}$ are sent into the human-box FFN,
and the $\boldsymbol{Q}_{O}^{out}$ are sent into the object-box and object-class FFN.
The final loss to be minimized is calculated in four parts:
location loss of human boxes ${L}^h_{loc}$,
that of object boxes ${L}^o_{loc}$,
object classification loss ${L}_{o}$,
and interaction classification loss ${L}_{a}$,
formulating as:
\begin{align}
    & {L}_{total} =
    {L}^h_{loc} +
    {L}^o_{loc} +
    \lambda_{o} {L}_{o} +
    \lambda_{a} {L}_{a},
\end{align}
where $\lambda_{o}$ and $\lambda_{a}$ are the weights of two classification losses, and ${L}_{loc}$ is computed by box regression ${L}_1$ loss and the GIoU loss~\cite{rezatofighi2019generalized} with weighting coefficients $\lambda_{reg}$ and $\lambda_{giou}$, which can be formulated as:
\begin{align}
    & {L}_{loc} =
    \lambda_{reg} {L}_{reg} +
    \lambda_{giou} {L}_{giou}.
\end{align}

The gradients w.r.t. ${L}^o_{loc}$ and ${L}_o$ update the parameters of the object decoder towards the optimal direction of object detection.
In general, the gradients w.r.t. ${L}_{a}$ will pass through both the human decoder and the object decoder since the interaction decoder receives the outputs from the two decoders.
However, considering the negative influence on object detection,
we stop the gradients w.r.t. ${L}_{a}$ propagating into the object decoder to keep optimal updates for object detection, which means the update of parameters of the object decoder is only computed by losses related to objects:
\begin{align}
    w_{o} \gets w_{o} - \alpha
    \left(
    \nabla_{w_{o}}
        \left(
            {L}^o_{loc} +
            \lambda_{o} {L}_{o}
        \right)
    \right),
\end{align}
where $w_{o}$ denotes the parameters of the object decoder and $\alpha$ is the learning rate.
For the learnable nature of the positional embeddings,
the update of the object queries $\boldsymbol{Q}_O$ can be represented as:
\begin{align}
    \boldsymbol{Q}_O \gets \boldsymbol{Q}_O - \alpha
    \left(
    \nabla_{\boldsymbol{Q}_O}
        \left(
            {L}^o_{loc} +
            \lambda_{o} {L}_{o}
        \right)
    \right).
\end{align}
Since human detection is not disturbed by interaction gradients but benefits from them,
we maintain the update for the human decoder with parameters $w_{h}$ as:
\begin{align}
    w_{h} \gets w_{h} - \alpha
    \left(
    \nabla_{w_{h}}
        \left(
            {L}^h_{loc} +
            \lambda_{a} {L}_{a}
        \right)
    \right).
\end{align}
So does the update for learnable human queries:
\begin{align}
    \boldsymbol{Q}_{h} \gets \boldsymbol{Q}_{h} - \alpha
    \left(
    \nabla_{\boldsymbol{Q}_{h}}
        \left(
            {L}^h_{loc} +
            \lambda_{a} {L}_{a}
        \right)
    \right).
\end{align}
With SG-Mechanism,
the detection of both humans and objects can achieve the best performance.
And with better detection performance,
the following interaction prediction can be further improved.

%% file: Figures/framework.tex
\begin{figure*}[t]
    \centering
    \includegraphics[width=1\textwidth]{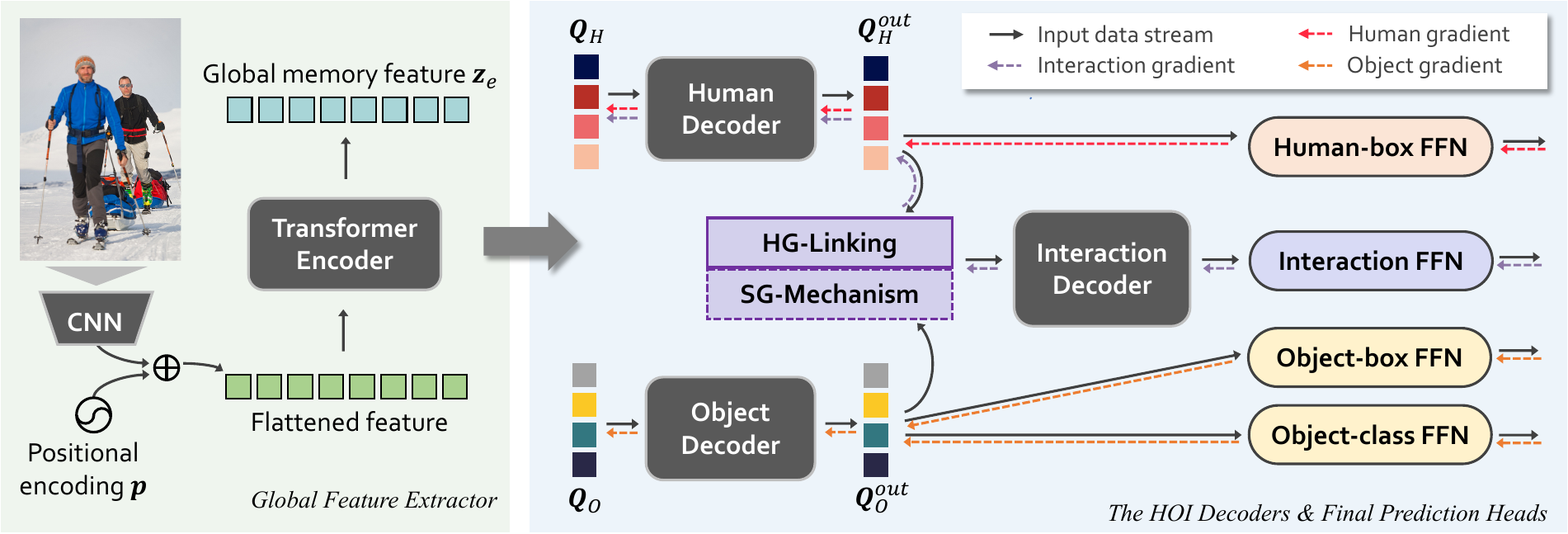}
    \caption{Overview framework of \textit{Human and Object Disentangling Network} (HODN).
    Given an input image, the global memory feature $\boldsymbol{z}_e$ is extracted by the CNN backbone and the transformer encoder.
    %
    Then two parallel decoders, \emph{i.e.}, human decoder and object decoder, introduce two sets of learnable positional embeddings ($\boldsymbol{Q}_H$ and $\boldsymbol{Q}_O$) to obtain human features and object features ($\boldsymbol{Q}_H^{out}$ and $\boldsymbol{Q}_O^{out}$).
    The interaction decoder receives them to mine interactive information by the Human-Guide Linking method (described in Section\,\ref{sec:met_on}). Finally, the outputs of three decoders are sent into corresponding FFNs to get HOI predictions. During the back-propagation stage of training, the \textit{Stop-Gradient Mechanism} (presented in Section\,\ref{sec:met_from}) is used to process the interaction gradients in a particular way.}
    \label{fig:framework}
\end{figure*}

%% file: Figures/dec_layer.tex
\begin{figure}[t]
    \centering
    \includegraphics[width=0.38\textwidth]{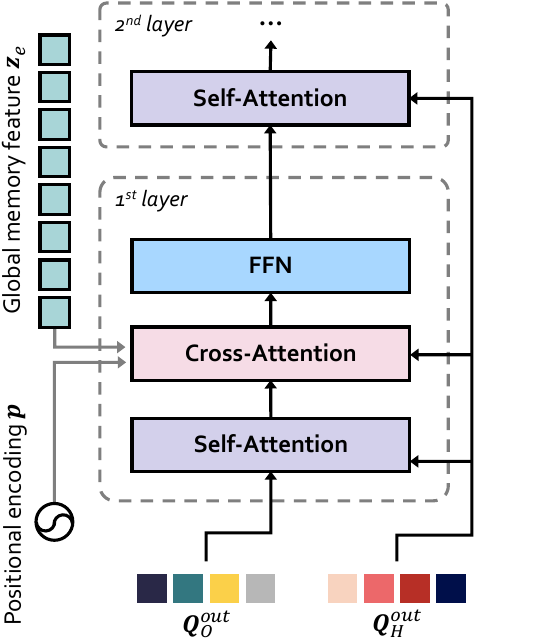}
    \caption{The details of the \textit{Human-Guide Linking} depicted in Section\,\ref{sec:met_on}.}
    \label{fig:dec_layer}
\end{figure}

%% file: sec_4_experiment.tex
\section{Experiment}
\input{Figures/motiv_mask.tex}

\subsection{Experiment Setup}
\subsubsection{Dataset}
We follow the evaluation of most previous works and report the mean average precision (mAP) on two public benchmarks, \emph{i.e.}, V-COCO~\cite{gupta2015visual} and HICO-Det~\cite{chao2018learning_two}.
V-COCO dataset, a subset of MS-COCO~\cite{lin2014microsoft}, contains 10,346 images (5,400 in the trainval set and 4,946 in the test set).
The images in it are annotated with 80 object classes and 29 action classes.
Among these action classes,
four of them are not associated with semantic roles,
so the role mAP of the V-COCO test set is only computed with the other 25 action classes.
In HICO-Det, there are 38,118 images for training and 9,658 for testing annotated with 80 object classes and 117 action classes.

\subsubsection{Metric}
We use the mean average precision (mAP) to report performance.
An HOI category is defined as an action class for V-COCO while a pair of an object class and an action class for HICO-Det.
Same as the standard evaluation scheme, a detection is judged as a true positive if the predicted human boxes and object boxes both have IoUs larger than 0.5 with the corresponding ground-truth boxes and if the predicted HOI category is also correct.
And the AP is calculated per HOI category.
For the V-COCO dataset, we report the role mAP in two scenarios, where scenario 1 needs to predict the cases in which humans interact with no objects while scenario 2 ignores these cases.
For the HICO-Det dataset, we report performance in two settings: default setting and known object setting.
In the former setting, the performance is evaluated on all test images while in the latter one, each AP is calculated on images that contain the target object class.
And in each setting, based on the number of HOI categories in the training set, we design three evaluation types:
full, rare, and non-rare.

\subsubsection{Implementation Details}
We adopt ResNet-50~\cite{he2016deep} followed by a six-layer transformer encoder as our global feature extractor.
For HOI decoders, the layer number of the human, object, and interaction decoder are all set to 6.
Layers in encoder and decoders have 8 heads, and the dimension inside the transformer architecture is 256.
The query size $N$ is set to 100 for V-COCO and 64 for HICO-Det since the average number of variant human-object pairs per image in V-COCO is larger than that in HICO-Det, which is as mentioned in CDN~\cite{zhang2021mining}.
The parameters of our proposed network are initialized with MS-COCO pre-trained DETR~\cite{carion2020end}.
For the missing parameters, we initialize them randomly.
During training, the AdamW~\cite{loshchilov2019decoupled} optimizer with the batch size of $16$, weight decay of ${10}^{-4}$, the initial learning rate of ${10}^{-5}$ for the backbone, and ${10}^{-4}$ for other parts is used.
For V-COCO, to eliminate overfitting,
we train our HODN for 90 epochs, with learning rates decaying by 10 times every 10 epochs after the 60th epoch and freeze the parameters of the backbone.
And for HICO-Det,
We train the whole HODN for 90 epochs where learning rates are decayed by 10 times at the 60th epoch.
The hyper-parameters weight coefficients in training loss ${\lambda}_{reg}, {\lambda}_{giou}, {\lambda}_o$ and ${\lambda}_a$ are set to 1, 2.5, 1, 1, respectively.
And the threshold of pair-wise NMS is set as 0.7 in inference, the same as CDN.
As for application experiments, all experimental settings are the same as those used in the original paper~\cite{zhang2022exploring_stip,liao2022gen_genvlkt} for a fair comparison.

\subsection{Relationships among humans, objects, and interactions}\label{sec:exp_motivation}
\input{Tables/object_mAP.tex}
In this sub-section, we first verify whether humans and objects have different mutual effects on the interactions.
In terms of \textit{how humans and objects impact interactions}, we argue that humans make more contributions to interaction prediction than objects since human feature contains more information (such as human postures and facial expression) that is strongly related to interactions~\cite{gkioxari2018detecting_two,bansal2020detecting,lin2021action_two}.
It can be easily verified by masking the regions of humans or objects in the images and inferring what the interaction is.
As in Figure\,\ref{fig:motiv_mask_obj},
although we can not see the object, it is still easy to infer that a man is throwing something.
On the contrary, in Figure\,\ref{fig:motiv_mask_human}, it is quite hard to determine whether the interaction is ``hit\_obj'' or ``catch\_obj'' or any else with only the ``sports ball'' visible.
To further verify the hypothesis, we use a pre-trained HOI detector, QPIC~\cite{tamura2021qpic}, to observe differences in results by masking humans and objects on the test set of V-COCO~\cite{gupta2015visual} respectively.
Note that the average of human areas and that of object areas are similar, which means that masking humans or objects may cause the same degree of information missing, so the performance comparison between the two is principally fair.
Results in Figure\,\ref{fig:motiv_mask_result} show that, without either humans or objects, the performance drops sharply.
Regardless of probability, the performance with masked humans is always lower than that with masked objects.
Furthermore, the performance gap becomes larger as the degree of masking increases.
That is to say, both of them make contributions to interaction prediction, but humans contribute much more than objects.

%
To further explore whether interactions impact humans and objects differently,
we modify QPIC by removing the action-class feed-forward network which is used to classify the interactions, and we train it with the same setting as QPIC.
The V-COCO benchmark serves as the training set and test set.
Without the interaction classification, all features are directly optimized for human detection and object detection so that the detection performance of both should be improved intuitively.
However, things do not turn out that way.
%
We report the results with and without interaction classification in Table\,\ref{tab:object_mAP}, along with the gaps as the subtraction of results without interaction and results with interaction.
As shown in Table\,\ref{tab:object_mAP}, without actions,
object detection can be improved a lot (recall rate, precision rate, and mAP are increased by 2.37, 3.03, and 1.65, respectively), implying that object detection is disturbed by actions.
On the contrary, the performance of human detection declines, with metrics decreasing by 0.21, 5.39, and 3.93, demonstrating the necessity for interactions.
We also conduct similar experiments on DisTR~\cite{zhou2022human_distr} and our HODN since both methods claim the disentanglement for instances and interactions.
%
From Table\,\ref{tab:object_mAP}, the slight performance discrepancies of object detection in DisTR indicate that separating detection from interactions is an efficient way to improve object detection.
The better results compared with QPIC verify this.
However, human detection in DisTR shows inadequate performance since the relatively independent instance detection stream stops interactions assisting human detection.
From the results of QPIC and DisTR, we can conclude that interactions influence human detection and object detection quite differently and instance-level disentanglement only helps object detection.
%
In terms of HODN, it shows significant superiority in both human and object detection.
The gap between HODN and HODN without action is also less than other methods,
we owe this to our disentangled human and object decoders and our SG-Mechanism strategy.

%
As analysis and experimental results above, we conclude the HOI relationships are:
1) for interaction prediction, human features make more contributions than object ones;
2) for human and object detection, interactive information assists in the former but obstructs the latter.
%

\input{Tables/hico.tex}
\input{Tables/vcoco.tex}
\subsection{Quantitative Analysis}
\subsubsection{Performance Comparisons}
We first evaluate the performance of our method on the HICO-Det test set with ResNet-50~\cite{he2016deep} as the backbone,
and report result in Table\,\ref{tab:hico}.
Our method achieves a competitive result, \emph{e.g.}, 33.14 mAP on the full evaluation for the default setting.
Compared with transformer-based single-decoder works HoiTransformer~\cite{zou2021end} and QPIC~\cite{tamura2021qpic},
our HODN has achieved 41.26\% and 14.00\% relative mAP gain.
Even when comparing PhraseHOI~\cite{li2022improving_transformer}, OCN~\cite{yuan2022detecting_transformer}, and SSRT~\cite{iftekhar2022look_transformer} which introduce prior knowledge of language,
our method still attains relative improvements of 13.14\%, 7.21\%, and 9.16\%.
We argue that this gap comes from the limitation of the single-decoder that can not explicitly model the HOI relationships regardless of adopting auxiliary knowledge.
Among multiple-decoders methods,
our method outperforms 32.03\% by HOTR~\cite{kim2021hotr}, 14.79\% by AS-Net~\cite{chen2021reformulating}, 4.28\% by CDN~\cite{zhang2021mining}, and 4.38\% by DisTR~\cite{zhou2022human_distr}.
Even with multiple decoders, these methods ignore the different potentiality between humans and objects, showing unsatisfactory performance.
We can conclude that except for the GEN-VLKT~\cite{liao2022gen_genvlkt} that introduces extra semantic information,
our method achieves the best performance.
We then conduct performance comparison experiments on the V-COCO test set.
From Table\,\ref{tab:vcoco},
our method achieves 67.0 role mAP in scenario 1 and 69.1 in scenario 2, outperforming all existing methods without extra knowledge.
In particular,
compared to methods only appearance feature referred to,
we promote the state-of-the-art work, DisTR~\cite{zhang2021mining}, with about 1.21\% and 0.88\% performance improvement under the two scenarios.
Moreover,
compared to methods with extra knowledge,
our method is still competitive with similar performance with state-of-the-art method STIP~\cite{zhang2022exploring_stip}
(67.0 of ours compared with 66.0 of STIP in scenario 1 and 69.1 compared with 70.7 in scenario 2).
%
Note that both GEN-VLKT and STIP introduce extra knowledge and complicated structures.
However, their performance varies on the two datasets.
GEN-VLKT introduces the text encoder in CLIP~\cite{radford2021learning} to initialize the weights of the interaction classifier FFN, which provides abundant prior knowledge on HICO-Det.
However, for V-COCO with a large number of categories, the training samples are insufficient to train which leads to inefficiency.
STIP uses graphs to construct the relationships among humans and objects. It is easy to form interactive relationships when dealing with a few HOI triplets.
Nevertheless, it becomes quite hard when the number of triplets increases.
Therefore, it performs excellent results on V-COCO, while showing insufficiency on dense triplets’ datasets like HICO-Det.
We argue that our method considers less extra knowledge which makes our method more generalized.
Our method performs consistently on both benchmarks with somehow similar performance to the best one on each dataset, which implies the practicality of our method.

\subsubsection{Application to Existing Works}
As we analyzed before, many existing works only disentangle HOI detection into instance detection and interaction prediction,
\emph{i.e.}, they only construct the relationships between instances and interactions, not distinguish humans and objects.
Note that our motivation and method are orthogonal to them.
We can combine HODN with them by further disentangling instances into humans and objects and adopting our proposed HG-Linking and SG-Mechanism.
Almost all two-decoder transformer-based methods can combine with ours.
Here we take STIP~\cite{zhang2022exploring_stip} and GEN-KLVT~\cite{liao2022gen_genvlkt} as examples,
which are state-of-the-art methods on V-COCO and HICO-Det, respectively.
%
As shown in the last row of Table\,\ref{tab:hico},
combined with our work,
the performance of GEN-VLKT increases by 2.40\% and 2.94\%, attaining 34.56 mAP and 37.86 mAP under default and known objects settings on HICO-Det.
As for STIP in the last row of Table\,\ref{tab:vcoco},
STIP can achieve 66.5 mAP and 71.5 role mAP V-COCO,
gaining improvements of 0.91\% and 1.13\%.
Both verify that existing methods can benefit easily from our method and achieve new state-of-the-art results.

\subsection{Ablation Study}
\subsubsection{HG-Linking}
The specialness of HG-Linking (Section\,\ref{sec:met_on}) is to hold one of the HOI relationships, \emph{i.e.},
making $\boldsymbol{Q}_H^{out}$ (human features) as the principal and serving $\boldsymbol{Q}_O^{out}$ (object features) as the auxiliary.
To prove its effectiveness,
we first verify the difference between the contributions of humans and objects.
Particularly,
we treat them as the same by passing the addition of $\boldsymbol{Q}_H^{out}$ and $\boldsymbol{Q}_O^{out}$ into the interaction decoder.
Here, we adopt two strategies to link:
1) the interaction decoder takes the addition as the positional embeddings (interaction queries) like a vanilla decoder;
2) the interaction decoder introduces a set of randomly initialized learnable positional embeddings and receives the addition as an extra input of the 1st self-attention module.
For convenience,
we name the first strategy as ``addition-guide'' and the second as ``random-guide''
and report the results in the first row and the second row of Table\,\ref{tab:ablation_part}, respectively.
We observe a performance decline of more than 1.3\% and 3.7\%, respectively,
indicating the necessity of disentangling the features of humans and objects.
The sharper performance drop of the ``random-guide'' compared to the ``addition-guide'' also indicates the non-trivial effect of the positional embeddings.
Then we verify the superior contribution of humans compared with objects by oppositely treating human and object features, \emph{i.e.}, ``object-guide''.
In the 3rd row of Table\,\ref{tab:ablation_part},
with ``object-guide'',
the performance decreases with a margin of 3.8 and 3.9 role mAP,
implying the more contributive potentiality of human features.
The degradation of performance with replaced link methods proves the effectiveness and verifies that humans and objects have different degrees of contributions to interactions, and the former are more helpful ones.

\input{Tables/ablation_part.tex}
\input{Figures/fvp.tex}

%
\subsubsection{SG-Mechanism}
\input{Figures/visualize.tex}
We further conduct the ablation study for SG-Mechanism (Section\,\ref{sec:met_from}).
We design SG-Mechanism by holding another HOI relationship:
interactions will aid human detection but interfere with object detection.
To check its effectiveness, we remove SG-Mechanism to allow interaction gradients to optimize the object decoder.
The performance gap is shown in the 4th row of Table\,\ref{tab:ablation_part} indicates that this mechanism does protect object detection from being disturbed by interactions.
As we stated, SG-Mechanism can eliminate the negative impact of interaction on objects, and better object detection can bring more accurate interaction prediction.
Moreover, we also apply SG-Mechanism to the detection of humans and show the result in row 5 of Table\,\ref{tab:ablation_part}.
Note that by adopting SG-Mechanism only for humans,
the interaction gradients will backpropagate to the object decoder while not to the human decoder,
which means not only the object decoder will be disturbed but also the human decoder can not benefit from interaction prediction.
The larger degradation of the role mAP (from 64.7 to 64.3 and from 66.9 to 66.6) implies that interactions assist in human detection though slightly.
We can conclude that
only by adopting SG-Mechanism to objects,
as our HODN does,
the detection of both humans and objects can achieve the best performance.

\subsection{Parameters vs. Performance}
Considering that the disentangling decoders introduce more parameters, we conduct experiments about the performance versus the number of parameters on the V-COCO test set.
Here we adopt two other smaller networks, named HODN-tiny (HODN-T) and HODN-small (HODN-S).
In particular,
two parameter-shared 3-layer decoders for humans and objects, as well as a 3-layer interaction decoder, are included in the HODN-T.
And two shared 6-layer decoders and one 3-layer interaction decoder are adopted in the HODN-S.
Note that we still utilize independent queries to disentangle the features of humans and objects even when parameters are shared.
The comparison result is shown in Figure\,\ref{fig:param} by reporting the role mAP under scenario 1 since some of the existing works do not support evaluation for scenario 2.
As shown in Figure\,\ref{fig:param},
HODN outperforms previous works under various settings of the number of parameters.
Even though the improvement compared with OCN~\cite{yuan2022detecting_transformer} is slight,
HODN-T introduces much fewer parameters (HODN-T with 39.8M and 41.8M with OCN).
The role mAP rises dramatically as the number of parameters increases and reaches an optimal performance at HODN setting.
Note that
even compared to CDN-L (63.9 mAP in scenario 1), which introduces much more parameters (about 67.0M), our HODN (with about 57.9M parameters) still maintains a significant advantage.
We conclude that there is no direct correlation between performance and the number of parameters.
Our efficiency comes more from our well-designed framework than from the larger number of parameters.

\input{Figures/attnmap.tex}

\subsection{Qualitative Results}
The performance of HOI detection relies on the accuracy of instance location and interaction prediction.
We argue that HODN can detect much more objects, especially occluded ones or overlapping ones.
We visualize some examples and use a classical HOI detector QPIC~\cite{tamura2021qpic} for comparison.
On one hand,
the location of objects will be disturbed if introducing interaction gradients and that of humans will be improved, as we previously examined.
As shown in the left-top images of Figure\,\ref{fig:visualize}, QPIC locates two ``surfboards'' in one box, while our HODN does not conflate a bunch of objects.
In the left-middle images, QPIC only locates a little part of the ``bed'', while our HODN can predict the entire ``bed'' even though most parts of it are occluded by a human.
Furthermore, in the left-bottom images,
even with only the legs visible,
HODN can still locate the man entirely with the help of interactions, whereas QPIC fails.
We owe this to SG-Mechanism, which assists in human detection and protects object detection from negative influence by interactions.
On the other hand,
with information like posture, human features are more vital to interaction prediction.
However,
object features, albeit helpful for interaction prediction, may introduce bias due to the imbalanced data distribution.
Therefore, the prediction of interactions is likely to overfit objects
if the network pays too much attention to object features.
For example, in the right-top images of Figure\,\ref{fig:attnmap},
QPIC mispredicts the interaction ``hold'' as ``ride'' when recognizing the ``motorcycle'', since ``A person is riding a motorcycle'' happens with a high probability.
A similar situation happens in the right-middle and the right-bottom images of Figure\,\ref{fig:attnmap},
where interaction ``lay'' has a high correlation with ``bed'' and ``using snowboard'' has a strong association with ``snowboard'', misleading the predictions of QPIC.
On the contrary, HODN predicts interactions accurately.
We owe this to HG-Linking, which makes the interaction decoder pay more attention to humans and less attention to objects.
We also visualize the attention maps extracted from the last layer of the decoder of QPIC
and decoders of HODN
in Figure\,\ref{fig:attnmap}.
To see more clearly, we also draw the ground truth bounding boxes of humans and objects.
Our three decoders can concentrate on their own goal regions, \emph{i.e.}, the human decoder pays attention to humans, the object decoder to objects, and the interaction decoder to the regions that contribute to understanding actions.
For example, in the left-bottom images, the human decoder focuses on the man's head and limbs, while the attention of the interaction decoder becomes more fine-grained after combining object features: the highlight parts become the man's hands and the sports ball.
In the right-bottom images, the interaction decoder shifts attention from the head and limbs of the man to the interaction regions where humans and horses come into contact.
On the contrary, QPIC uses only one decoder to handle human detection, object detection, and interaction prediction.
Consequently, it can not distinguish their difference.
As in the visualization of attention maps, QPIC only focuses on object-around regions the most of time.
Therefore, the missing objects or incorrect location may cause QPIC to fail to find the correct HOI triplets.
This also explains why our method performs much better than it.


%% file: Figures/motiv_mask.tex
\begin{figure}[t]
    \centering
    \subfloat[Masking objects]{
        \includegraphics[width=0.225\textwidth]{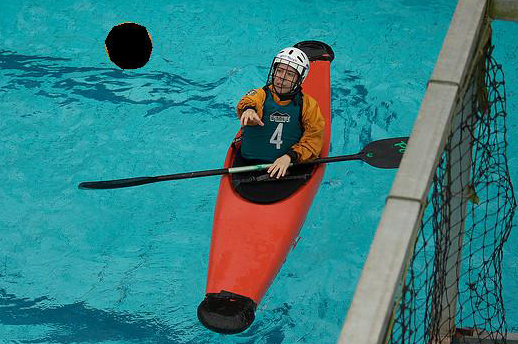}
        \label{fig:motiv_mask_obj}
    }
    \subfloat[Masking humans]{
       \includegraphics[width=0.225\textwidth]{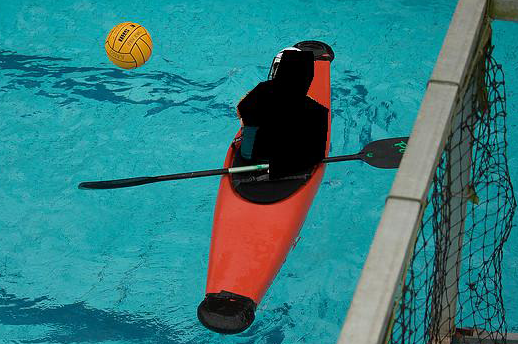}
       \label{fig:motiv_mask_human}
    }
    \\
    \subfloat[Performance vs. Mask-degree]{
       \includegraphics[width=0.47\textwidth]{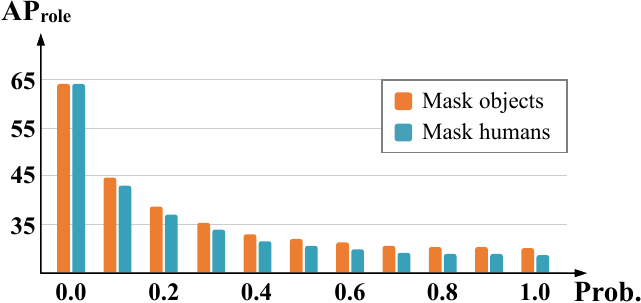}
       \label{fig:motiv_mask_result}
    }
    \caption{The interaction category of both (a) and (b) is ``throw\_obj''. For human understanding, it is easy to determine the interaction class in (a) but hard in (b). And in (c), the mAP results based on the V-COCO test set demonstrate that the HOI detection with masked humans is much worse than that with masked objects and the gap becomes larger with higher masking probability.}
\end{figure}

%% file: Tables/object_mAP.tex
\begin{table}[t]
    \centering

    \caption{Comparison of human and object detection performance on V-COCO test set with or without action branch by QPIC~\cite{tamura2021qpic}, DisTR~\cite{zhou2022human_distr}, and our HODN. \textbf{R} and \textbf{P} represent Recall Rate and Precision Rate relatively. The subscripts \textbf{o} and \textbf{h} denote objects and humans respectively. The ``gap'' row is calculated as the subtraction of results without and with action.}

    \label{tab:object_mAP}

    \resizebox{0.48\textwidth}{!}{
        \begin{tabular}{lllllll}
        \toprule[1pt] &
        $\mathrm{R_o}$ & $\mathrm{P_o}$ & $\mathrm{mAP_o}$ &
        $\mathrm{R_h}$ & $\mathrm{P_H}$ & $\mathrm{AP_h}$ \\
        \midrule

        \textbf{QPIC} &
        73.72    & 21.51    &  43.69 &
        98.67    &  6.92    &  74.95 \\

        w/o action  &
        76.10 ($\uparrow$)   & 24.54 ($\uparrow$)   &  45.34 ($\uparrow$) &
        98.46 ($\downarrow$) &  1.53 ($\downarrow$) &  71.02 ($\downarrow$) \\

        gap &
        2.38 & 3.03 & 1.65 & -0.21 & -5.39 & -3.93 \\

        \hdashline[3pt/3pt]\rule{-3pt}{10pt}

        \textbf{DisTR} &
        77.11    & 23.51    &  44.84 &
        98.64    &  4.99    &  73.80 \\

        w/o action  &
        78.03 ($\uparrow$)   & 23.72 ($\uparrow$)   & 45.20 ($\uparrow$) &
        98.44 ($\downarrow$) &  4.00 ($\downarrow$) & 72.67 ($\downarrow$) \\

        gap &
        0.92 & 0.21 & 0.36 & -0.20 & -0.99 & -1.13 \\

        \hdashline[3pt/3pt]\rule{-3pt}{10pt}

        \textbf{HODN} &
        85.20    & 25.10    &  53.20 &
        98.93    &  5.77    &  81.68 \\

        w/o action  &
        86.01 ($\uparrow$)   & 25.28 ($\uparrow$)   &  53.31 ($\uparrow$) &
        98.78 ($\downarrow$) &  5.04 ($\downarrow$) &  80.59 ($\downarrow$) \\

        gap &
        0.81 & 0.18 & 0.11 & -0.15 & -0.73 & -1.09 \\

        \bottomrule[1pt]
        \end{tabular}
    }
\end{table}

%% file: Tables/hico.tex
\begin{table*}[t]
\centering

\caption{Performance comparison on HICO-Det dataset.
Each letter in the Feature column stands for \textbf{A}: Appearance/Visual feature, \textbf{S}: Spatial features~\cite{gao2018ican_two}, \textbf{L}: Linguistic feature of label semantic embeddings, \textbf{P}: Human pose feature.
}

\label{tab:hico}

\begin{tabular}{
    p{80pt}lp{50pt}<{\centering}
    cccccc
}
\toprule[1pt]
\multicolumn{1}{l}{} & \multicolumn{1}{l}{} & \multicolumn{1}{l}{} & \multicolumn{3}{c}{\textbf{Default}} &
\multicolumn{3}{c} {\textbf{Known Object}} \\ [2pt]
\textbf{Method} & Backbone & Feature &
Full & Rare & NonRare      &
Full & Rare & NonRare      \\ \midrule

\multicolumn{9}{l}{\textbf{\textit{Traditional Methods:}}} \\
iCAN~\cite{gao2018ican_two}
    & ResNet-50
    & A+S
    & 14.84 & 10.45 & 16.15 & 16.26 & 11.33 & 17.73 \\
iHOI~\cite{xu2019interact_two_pose}
    & ResNet-50-FPN
    & A+S
    & 13.39 &  9.51 & 14.55 & -     & -     & -     \\
TIN~\cite{li2019transferable_two}
    & ResNet-50
    & A+S+P
    & 17.22 & 13.51 & 18.32 & 19.38 & 15.38 & 20.57 \\
DRG~\cite{gao2020drg_two}
    & ResNet-50-FPN
    & A+S+P+L
    & 19.26 & 17.74 & 19.71 & 23.40 & 21.75 & 23.89 \\
VCL~\cite{hou2020visual_two}
    & ResNet-50
    & A+S
    & 19.43 & 16.55 & 20.29 & 22.00 & 19.09 & 22.87 \\
VSGNet~\cite{hou2020visual_two}
    & ResNet-152
    & A+S
    & 19.80 & 16.05 & 20.91 & -     & -     & -     \\
FCMNet~\cite{liu2020amplifying_two}
    & ResNet-50
    & A+S+P
    & 20.41 & 17.34 & 21.56 & 22.04 & 18.97 & 23.12 \\
ACP~\cite{kim2020detecting_two}
    & ResNet-152
    & A+S+P
    & 20.59 & 15.92 & 21.98 & -     & -     & -     \\
PastaNet~\cite{li2020pastanet_two}
    & ResNet-50
    & A+P
    & 22.65 & 21.17 & 23.09 & 24.53 & 23.00 & 24.99 \\
ConsNet~\cite{liu2020consnet_two}
    & ResNet-50-FPN
    & A+S+L
    & 22.15 & 17.12 & 23.65 & - & - & - \\
IDN~\cite{li2020hoi_two}
    & ResNet-50
    & A+S
    & 23.36 & 22.47 & 23.63 & 26.43 & 25.01 & 26.85 \\
UnionDet~\cite{kim2020uniondet_one}
    & ResNet-50-FPN
    & A
    & 17.58 & 11.72 & 19.33 & 19.76 & 14.68 & 21.27 \\
IP-Net~\cite{wang2020learning_one}
    & Hourglass-104
    & A
    & 19.56 & 12.79 & 21.58 & 22.05 & 15.77 & 23.92 \\
PPDM~\cite{liao2020ppdm_one}
    & Hourglass-104
    & A
    & 21.73 & 13.78 & 24.10 & 24.58 & 16.65 & 26.84 \\
GG-Net~\cite{zhong2021glance_one}
    & Hourglass-104
    & A
    & 23.47 & 16.48 & 25.60 & 27.36 & 20.23 & 29.48 \\
\midrule

\multicolumn{9}{l}{\textbf{\textit{Transformer-based Methods:}}} \\
HoiTransformer~\cite{zou2021end}
    & ResNet-50
    & A
    & 23.46 & 16.91 & 25.41 & 26.15 & 19.24 & 28.22 \\
HOTR~\cite{kim2021hotr}
    & ResNet-50
    & A
    & 25.10 & 17.34 & 27.42 & -     & -     & -     \\
AS-Net~\cite{chen2021reformulating}
    & ResNet-50
    & A
    & 28.87 & 24.25 & 30.25 & 31.74 & 27.07 & 33.14 \\
QPIC~\cite{tamura2021qpic}
    & ResNet-50
    & A
    & 29.07 & 21.85 & 31.23 & 31.68 & 24.14 & 33.93 \\
CDN~\cite{zhang2021mining}
    & ResNet-50
    & A
    & 31.78 & 27.55 & 33.05 & 34.53 & 29.73 & 35.96 \\
PhraseHOI~\cite{li2022improving_transformer}
    & ResNet-50
    & A+L
    & 29.29 & 22.03 & 31.46 & 31.97 & 23.99 & 34.36 \\
OCN~\cite{yuan2022detecting_transformer}
    & ResNet-50
    & A+L
    & 30.91 & 25.56 & 32.51 & - & - & - \\
SSRT~\cite{iftekhar2022look_transformer}
    & ResNet-50
    & A+L
    & 30.36 & 25.42 & 31.83 & - & - & - \\
DisTR~\cite{zhou2022human_distr}
    & ResNet-50
    & A
    & 31.75 & 27.45 & 33.03 & 34.50 & 30.13 & 35.81 \\
STIP~\cite{zhang2022exploring_stip}
    & ResNet-50
    & A+S+L
    & 32.22 & 28.15 & 33.43 & 35.29 & 31.43 & 36.45 \\
GEN-VLKT~\cite{liao2022gen_genvlkt}
    & ResNet-50
    & A+L
    & 33.75 & 29.25 & 35.10
    & 36.78 & 32.75 & 37.99 \\
\hdashline[3pt/3pt]\rule{-3pt}{10pt}
HODN
    & ResNet-50
    & A
    & 33.14 & 28.54 & 34.52
    & 35.86 & 31.18 & 37.26 \\
HODN + GEN-VLKT
    & ResNet-50
    & A+L
    & \textbf{34.56} & \textbf{30.26} & \textbf{35.84}
    & \textbf{37.86} & \textbf{33.93} & \textbf{39.03} \\

\bottomrule[1pt]
\end{tabular}

\end{table*}

%% file: Tables/vcoco.tex
\begin{table}[t!]
\centering
\caption{Performance comparison on V-COCO dataset. ${\mathrm{AP}}_{\mathrm{role}}^{\#1}$ and ${\mathrm{AP}}_{\mathrm{role}}^{\#2}$ are role mAP in scenario 1 and 2.
In the Backbone column, \textbf{R} and \textbf{HOG} represent ResNet and Hourglass, respectively.
Each letter in the Feature column stands for \textbf{A}: Appearance/Visual feature, \textbf{S}: Spatial features~\cite{gao2018ican_two}, \textbf{L}: Linguistic feature of label semantic embeddings, \textbf{P}: Human pose feature.
}

\label{tab:vcoco}

\begin{tabular}{
    p{70pt}lccc
}
\toprule[1pt]
\textbf{Method}  & Backbone  & Feature &
${\mathrm{AP}}_{\mathrm{role}}^{\#1}$ &
${\mathrm{AP}}_{\mathrm{role}}^{\#2}$ \\ \midrule

\multicolumn{4}{l}{\textbf{\textit{Traditional Methods:}}} \\
iCAN~\cite{gao2018ican_two} & R50 & A+S &
45.3      & 52.4 \\
iHOI~\cite{xu2019interact_two_pose} & R50-FPN & A+P &
45.8      & -    \\
TIN~\cite{li2019transferable_two} & R50 & A+S+P &
48.7      & -    \\
DRG~\cite{gao2020drg_two} & R50 & A+S+P+L &
51.0      & -    \\
VCL~\cite{hou2020visual_two} & R50 & A+S &
48.3      & -    \\
VSGNet~\cite{hou2020visual_two} & R152 & A+S &
51.8      & -    \\
FCMNet~\cite{liu2020amplifying_two} & R50 & A+S+P &
53.1      & -    \\
ACP~\cite{kim2020detecting_two} & R152 & A+S+P &
53.2      & -    \\
PastaNet~\cite{li2020pastanet_two} & R50 & A+P &
51.0      & 57.5 \\
ConsNet~\cite{liu2020consnet_two} & R50-FPN & A+S+L &
53.2      & -    \\
IDN~\cite{li2020hoi_two} & R50 & A+S &
53.3      & 60.3 \\
UnionDet~\cite{kim2020uniondet_one} & R50-FPN & A &
47.5      & 56.2 \\
IP-Net~\cite{wang2020learning_one} & HOG104 & A &
51.0      & -    \\
PPDM~\cite{liao2020ppdm_one} & HOG104 & A &
-         & -    \\
GG-Net~\cite{zhong2021glance_one} & HOG104 & A &
54.7      & -    \\ \midrule

\multicolumn{4}{l}{\textbf{\textit{Transformer-based Methods:}}} \\
HoiTransformer~\cite{zou2021end} & R50 & A &
52.9      & -    \\
HOTR~\cite{kim2021hotr} & R50 & A &
55.2      & 64.4 \\
AS-Net~\cite{chen2021reformulating} & R50 & A &
53.9      & -    \\
QPIC~\cite{tamura2021qpic} & R50 & A &
58.8      & 61.0 \\
CDN~\cite{zhang2021mining} & R50 & A &
62.3      & 64.4 \\
PhraseHOI~\cite{li2022improving_transformer} & R50 & A+L &
57.4      & - \\
OCN~\cite{yuan2022detecting_transformer} & R50 & A+L &
64.2      & 66.3 \\
SSRT~\cite{iftekhar2022look_transformer} & R50 & A+L &
63.7      & 65.9 \\
DisTR~\cite{zhou2022human_distr} & R50 & A &
66.2      & 68.5 \\
STIP~\cite{zhang2022exploring_stip} & R50 & A+S+L &
66.0      & 70.7 \\
GEN-VLKT~\cite{liao2022gen_genvlkt} & R50 & A+L &
62.4      & 64.5 \\
\hdashline[3pt/3pt]\rule{-3pt}{10pt}
HODN      & R50 & A &
67.0      & 69.1 \\
HODN+STIP & R50 & A+S+L &
\textbf{67.5}      & \textbf{71.9} \\

\bottomrule[1pt]
\end{tabular}

\end{table}

%% file: Tables/ablation_part.tex
\begin{table}[t]
\centering

\caption{
    Ablation experiments on V-COCO test set,
    where row 1 to 3 are ablation studies for HG-Linking introduced in Section\,\ref{sec:met_on}, and row 4 to 5 for SG-Mechanism depicted in Section\,\ref{sec:met_from}.
    For HG-Linking, we compare different strategies to link decoders, which includes taking the addition of human features and object features (row 1), random learnable vectors (row 2), and object features (row 3) as positional embeddings of the interaction decoder, respectively.
    For SG-Mechanism,
    ablation experiments involve SG-Mechanism removal (row 4) and SG-Mechanism adoption to the human decoder (row 5).
}

\label{tab:ablation_part}

\begin{tabular}{clll}
\toprule[1pt]
\#Row & Ablation Item
& ${\mathrm{AP}}_{\mathrm{role}}^{\#1}$
& ${\mathrm{AP}}_{\mathrm{role}}^{\#2}$ \\ \midrule
  &\textbf{HODN}
    & \textbf{67.0}            & \textbf{69.1}            \\
\hdashline[3pt/3pt]\rule{-3pt}{10pt}
1 &human-guide $\rightarrow$ addition-guide
    & 65.1 ($\downarrow$ 1.9) & 67.2 ($\downarrow$ 1.9) \\
2 &human-guide $\rightarrow$ random-guide
    & 62.7 ($\downarrow$ 4.4) & 64.8 ($\downarrow$ 4.3) \\
3 &human-guide $\rightarrow$ object-guide
    & 63.3 ($\downarrow$ 3.8) & 65.2 ($\downarrow$ 3.9) \\
\hdashline[3pt/3pt]\rule{-3pt}{10pt}
4 &w/o SG-Mechanism
    & 64.7 ($\downarrow$ 2.3) & 66.9 ($\downarrow$ 2.2) \\
5 &SG-Mechanism to Human
    & 64.3 ($\downarrow$ 2.7) & 66.6 ($\downarrow$ 2.5) \\

\bottomrule[1pt]
\end{tabular}

\end{table}

%% file: Figures/fvp.tex
\begin{figure}[t]
    \centering
    \includegraphics[width=0.95\linewidth]{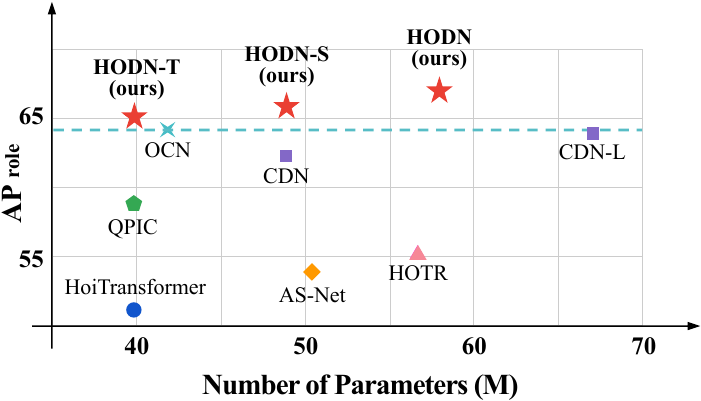}
    \caption{Performance comparison based on the different number of parameters.
    All results are conducted under scenario 1 on the V-COCO test set.
    The light blue dashed line, representing the performance of the state-of-the-art, OCN~\cite{yuan2022detecting_transformer}, with the role mAP of 64.2, is viewed as the baseline to measure against.
    Our HODN with tiny, small, and base parameter settings is signified by three red stars.
    And the three red stars locate higher than the light blue dashed line, demonstrating all HODNs outperform previous methods.
    }
    \label{fig:param}
\end{figure}



%% file: Figures/visualize.tex
\begin{figure*}[t]
    \centering
    \includegraphics[width=1\textwidth]{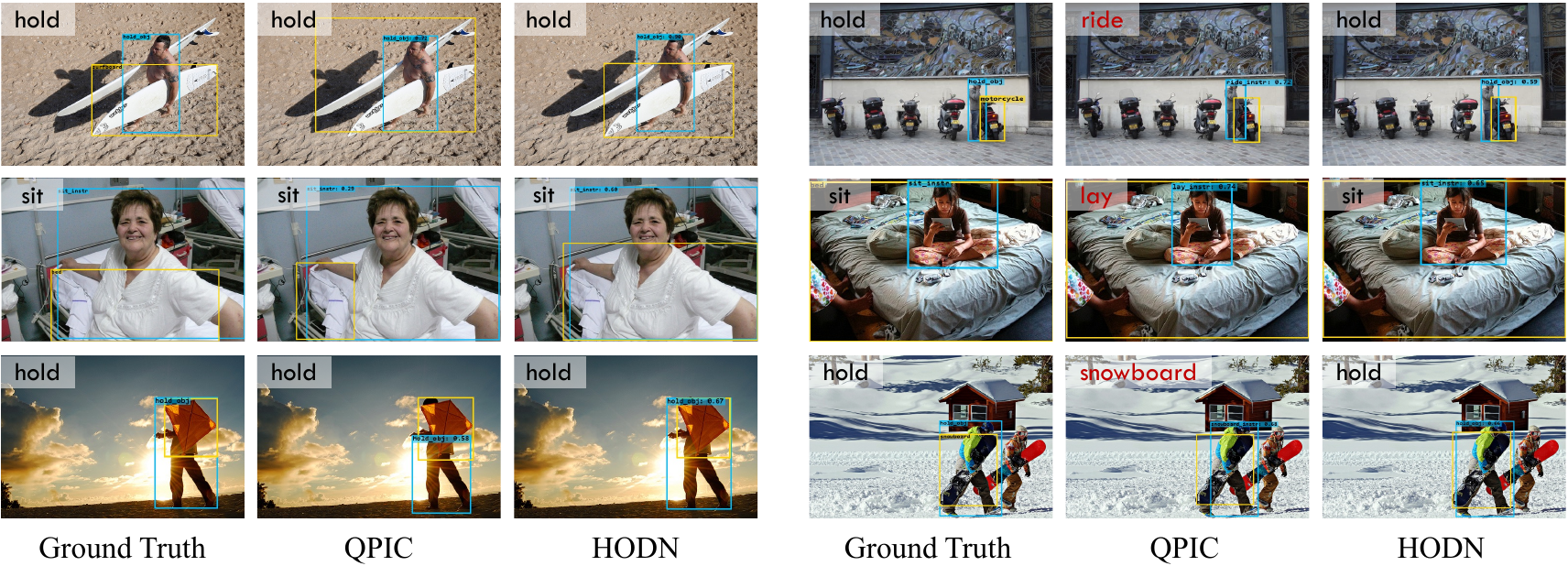}

    \caption{Visualization samples with ground-truth (left), as well as the detection results of QPIC (middle) and HODN (right). Bounding boxes of humans and objects are drawn with blue and yellow boxes. Interaction categories with confidence are depicted with blue characters. For clearer visualization, we zoom in interaction categories to the left-top corner of images with black characters for correct predictions and red for incorrect ones.}
    \label{fig:visualize}
\end{figure*}

%% file: Figures/attnmap.tex
\begin{figure*}[th]
    \centering
    \hspace{-15pt}
    \includegraphics[width=1.0\textwidth]{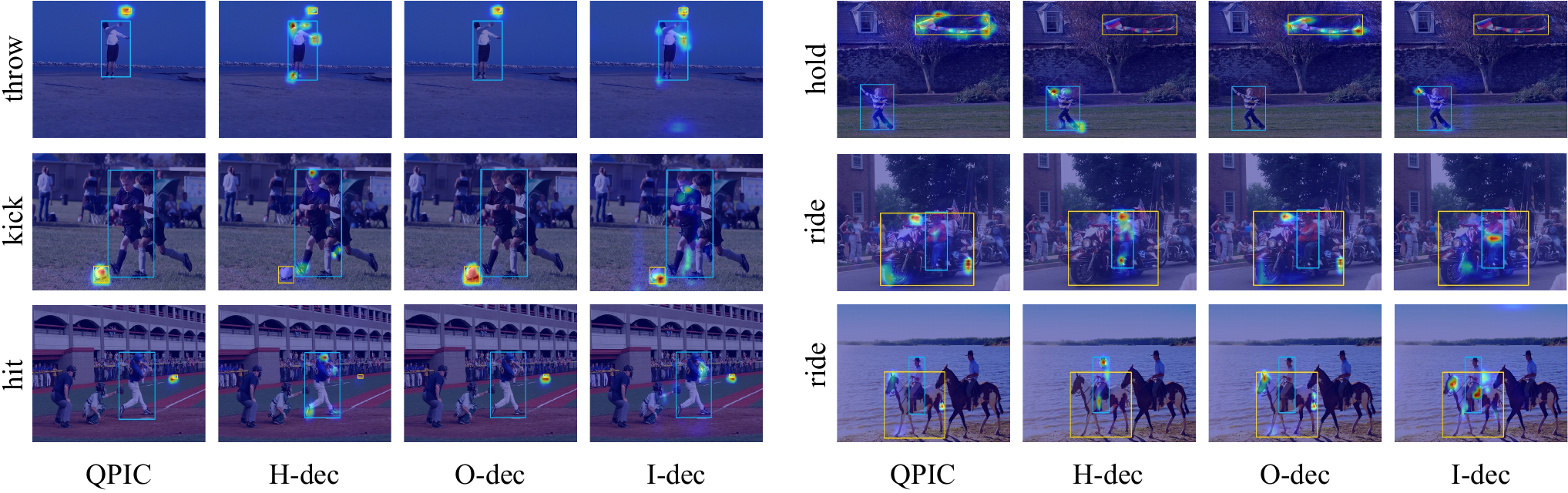}

    \caption{Visualization of attention maps of decoders in QPIC and HODN with the corresponding ground-truth bounding boxes. The interaction categories are depicted on the left side of the images. As can be seen from the figure, decoders in HODN can capture finer areas according to targets, while QPIC mixes all of them and is biased toward object regions. H-dec: the human decoder, O-dec: the object decoder, I-dec: the interaction decoder.}
    \label{fig:attnmap}
\end{figure*}

%% file: sec_5_conclusion.tex
\section{Conclusion}
In this paper, we analyze the relationships among humans, objects, and interactions in two aspects:
1) for interaction, human features make more contributions;
2) for detection, interactive information helps human detection while disturbing object detection.
Accordingly,
we propose a \textit{Human and Object Disentangling Network} (HODN), a transformer-based framework to explicitly model the relationships, which contains two parallel detection decoders for human and object features, and one interaction decoder for final interactions.
A \textit{Human-Guide Linking} method is used by the interaction decoder to make human features dominant and object ones auxiliary.
Particularly,
human features are sent into the interaction decoder as positional embeddings to make the decoder focus on human-centric regions.
Finally, considering that interactive information has the opposite influence on human detection and object detection,
we propose \textit{Stop-Gradient Mechanism}, where interaction gradients are not utilized to optimize object detection but human detection.
Since our method is orthogonal to the existing methods, they can be easily combined with our method and benefit from it.
Extensive experiments conducted on V-COCO and HICO-DET demonstrate that our method brings a significant performance improvement over the state-of-the-art HOI detection methods.

%% file: acknowledgements.tex
\section*{Acknowledgements}

This work was supported by National Key R\&D Program of China (No.2022ZD0118202), the National Science Fund for Distinguished Young Scholars (No.62025603), the National Natural Science Foundation of China (No. U21B2037, No. U22B2051, No. 62176222, No. 62176223, No. 62176226, No. 62072386, No. 62072387, No. 62072389, No. 62002305 and No. 62272401), and the Natural Science Foundation of Fujian Province of China (No.2021J01002,  No.2022J06001).